\pdfoutput=1

\documentclass[11pt]{article}

\usepackage{acl}

\usepackage{times}
\usepackage{latexsym}

\usepackage[T1]{fontenc}

\usepackage[utf8]{inputenc}

\usepackage{microtype}

\usepackage{mathtools}
\usepackage{bm}
\usepackage{soul}
\usepackage{caption}
\usepackage{subcaption}
\usepackage{array,multirow,graphicx}
\usepackage{floatrow}
\newfloatcommand{capbtabbox}{table}[][\FBwidth]
\usepackage{wrapfig,lipsum,booktabs}
\usepackage[flushleft]{threeparttable}
\usepackage{tabularx}
\usepackage{footnote}
\usepackage{siunitx}
\usepackage[export]{adjustbox} %
\usepackage{amssymb}%
\usepackage{pifont}%
\newcommand{\cmark}{\ding{51}}
\newcommand{\xmark}{\ding{55}}

\usepackage{ulem}
\usepackage{makecell}
\title{Progressive Class Semantic Matching for\\ Semi-supervised Text Classification}

\author{Hai-Ming Xu \and Lingqiao Liu \and Ehsan Abbasnejad \\
        Australian Institute for Machine Learning \\ The University of Adelaide \\
Adelaide, Australia \\
\texttt{\{hai-ming.xu, lingqiao.liu, ehsan.abbasnejad\}@adelaide.edu.au}}

\begin{document}
\maketitle
\begin{abstract}
Semi-supervised learning is a promising way to reduce the annotation cost for text-classification. Combining with pre-trained language models (PLMs), e.g., BERT, recent semi-supervised learning methods achieved impressive performance. In this work, we further investigate the marriage between semi-supervised learning and a pre-trained language model. Unlike existing approaches that utilize PLMs only for model parameter initialization, we explore the inherent topic matching capability inside PLMs for building a more powerful semi-supervised learning approach. Specifically, we propose a joint semi-supervised learning process that can progressively build a standard $K$-way classifier and a matching network for the input text and the Class Semantic Representation (CSR). The CSR will be initialized from the given labeled sentences and progressively updated through the training process. By means of extensive experiments, we show that our method can not only bring remarkable improvement to baselines, but also overall be more stable, and achieves state-of-the-art performance in semi-supervised text classification. Code is available at:~\url{https://github.com/HeimingX/PCM}.
\end{abstract}

\section{Introduction}
Text classification is a fundamental task in natural language processing (NLP) and underpins various applications, e.g., spam detection~\cite{jindal2007review}, sentiment analysis~\cite{pang2002thumbs} and text summarization~\cite{gambhir2017recent}. Supervised training of text classifiers often demands a large amount of annotation, which can be expensive for many applications. Semi-supervised learning (SSL) provides an economical way for alleviating this burden since it can make use of easy-accessible unlabeled samples to build a reasonably performed classifier with a limited amount of labeled data. Recently, SSL received increasing attention in both image classification~\cite{tarvainen2017mean,berthelot2019mixmatch,NEURIPS2020_06964dce} and text classification~\cite{xie2019unsupervised,chen2020mixtext,liu2021flitext} areas.

Meanwhile, pre-trained language models (PLMs)~\cite{yang2019xlnet,devlin2018bert,radford2019language} are developing rapidly and achieve impressive performance in various NLP tasks~\cite{sun2019utilizing, zhu2020incorporating} including text classification~\cite{garg2020bae}. In the context of semi-supervised text classification, many existing methods achieve excellent performance by directly using a PLM as a sentence encoder and further fine-tuning it with a semi-supervised learning process~\cite{xie2019unsupervised,chen2020mixtext,bhattacharjee2020bert,sun2020neural}. 
\begin{figure*}[t]
    \centering
    \begin{subfigure}[b]{0.8\textwidth}
        \centering
        \includegraphics[width=1.0\textwidth]{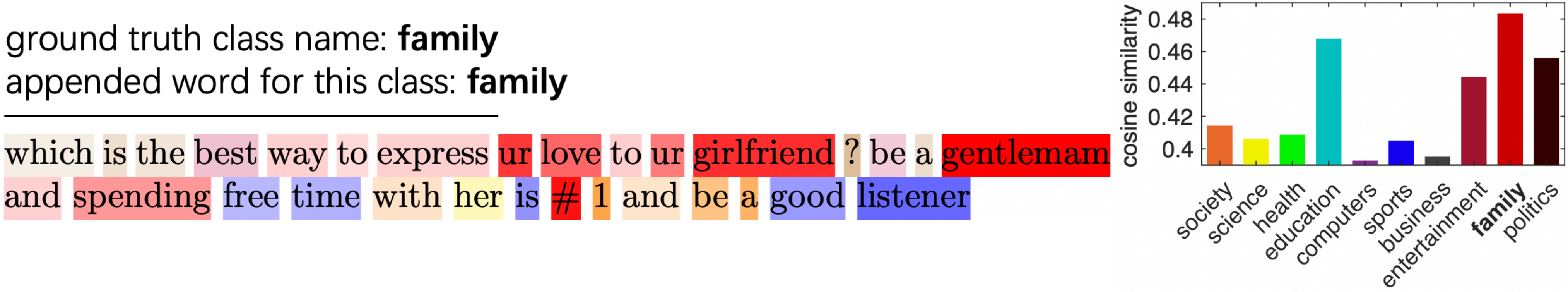}
        \includegraphics[width=1.0\textwidth]{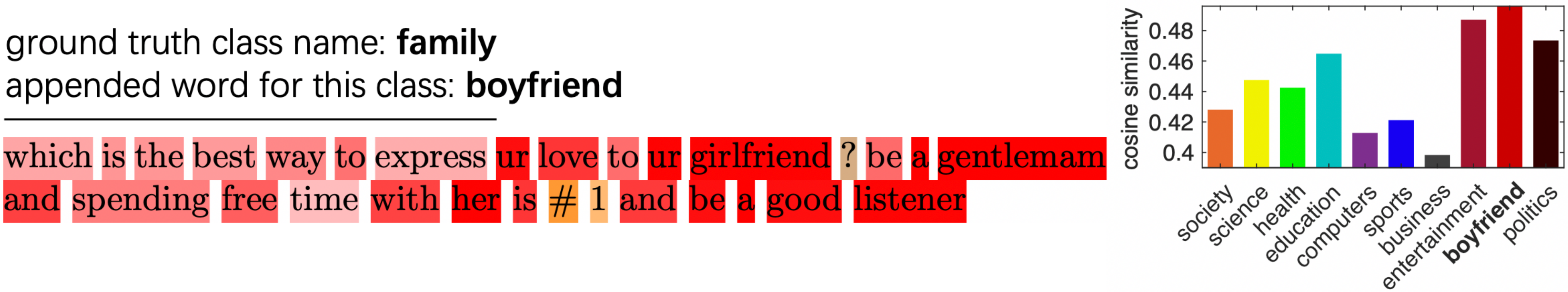}
        \caption{Sentence example on class ``family''}
        \label{fig:motivation_family}
    \end{subfigure}
    \hfill
    \begin{subfigure}[b]{0.8\textwidth}
        \centering
        \includegraphics[width=1.0\textwidth]{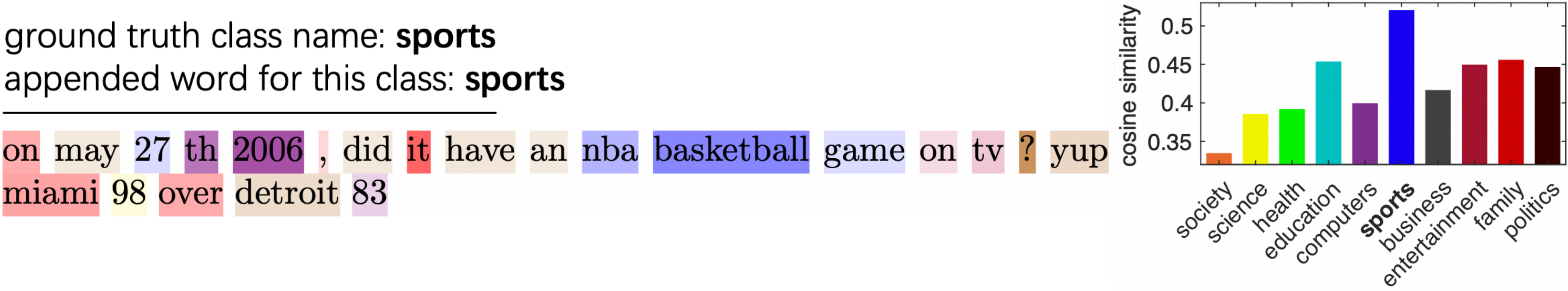}
        \includegraphics[width=1.0\textwidth]{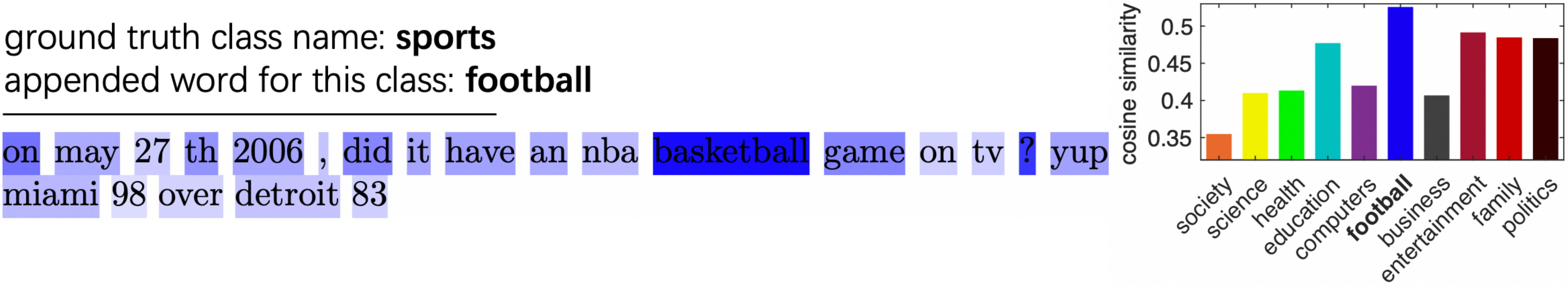}
        \caption{Sentence example on class ``sports''}
        \label{fig:motivation_sports}
    \end{subfigure}
    \centering
    \caption{Visualization of the inherent matching capability of BERT on examples from Yahoo! Answers. We append class semantic-related words (CSW) of all classes at the end of input sentence
    . Different colors denote different classes. 
    The color on each token of input sentence represents the category of its most attended CSW (with color brightness indicating the attention value, please see Sec.3 for more details). 
    The histograms on the right demonstrate the cosine similarity between the average features of sentence and features of each CSW.}
    \label{fig:motivation}
\end{figure*}

In this paper, we further explore the usage of PLMs for SSL. We go beyond the strategy of using PLMs for encoder initialization and make full use of inner knowledge of PLMs. Concretely, we identify that some PLMs, e.g., BERT, have an inherent matching capability between sentence and class-related words thanks to its pre-training pretext task~\cite{devlin2018bert} (as the examples shown in Fig.~\ref{fig:motivation}). We further propose to strengthen this capability through SSL on labeled and unlabeled data. Specifically, we develop a joint training process to update three components progressively, that is, a classifier that performs the standard $K$-way classification, a class semantic representation (CSR) that represents the semantic of each category, and a matching classifier that matches the input sentence against the CSR. Those three components can help each other during the training process, i.e., the $K$-way classifier will receive more accurate pseudo-labels by jointly generating pseudo-labels with the matching classifier; the matching classifier will also upgrade its matching capability with the guidance of the $K$-way classifier. The CSR will become more accurate and comprehensive with the improvement of the $K$-way classifier and matching classifier. This joint process leads to a more powerful semi-supervised learning algorithm for the text classification task. Throughout our experimental evaluation, we demonstrate that the proposed method achieves the state-of-the-art performance on text data, especially when the number of labeled sentences becomes extremely low, i.e., 3 or 5. 

%

%

%

\section{Related work}
In this section, we briefly review the relevant research works.

\subsection{General Semi-Supervised Learning}
Semi-supervised learning is a longstanding research topic in machine learning. Existing methods adopt different ways of utilizing unlabeled samples, e.g., ``transductive'' models~\cite{joachims2003transductive,gammerman2013learning}, multi-view style approaches~\cite{blum1998combining,zhou2005tri} and generative model-based methods~\cite{kingma2014semi,springenberg2015unsupervised}. With the renaissance of the deep neural network, consistency-regularization-based deep SSL approaches~\cite{laine2016temporal,tarvainen2017mean,miyato2018virtual} have achieved impressive performance on various tasks, and our work largely builds upon the method in this category. The key idea of these methods is to constrain the model to be consistent in the neighborhood of each sample in the input space. Specifically, $\Pi$-Model~\cite{laine2016temporal}, UDA~\cite{xie2019unsupervised} and FixMatch~\cite{NEURIPS2020_06964dce} directly add various perturbations to the input data, Mean-teacher~\cite{tarvainen2017mean} uses a teacher model to simulate sample perturbation, and Virtual Adversarial Training~\cite{miyato2018virtual} skillfully constructs an adversarial sample. More recently, mixup~\cite{zhang2017mixup} method proposed another kind of consistency constraint that requires the input and output of the model to satisfy an identical linear relationship. Based on this technique, many state-of-the-art methods are published, e.g., ICT~\cite{verma2019interpolation}, MixMatch~\cite{berthelot2019mixmatch} and ReMixMatch~\cite{berthelot2019remixmatch}.

\subsection{Semi-Supervised Text Classification}
Semi-supervised learning has gained a lot of attention in the field of text classification. Many recent semi-supervised text classification methods focus on how to adapt the existing SSL methodologies to the sentence input. \cite{miyato2016adversarial} applied perturbations to word embeddings for constructing adversarial and virtual adversarial training. \cite{clark2018semi} designed auxiliary prediction modules with restricted views of the input to encourage consistency across views. With the development of PLMs, \cite{jo2019delta} performed self-training between two sets of classifiers which are initialized differently, one with pre-trained word embeddings and random values for the other. Both~\cite{xie2019unsupervised} and~\cite{chen2020mixtext} took the pre-trained BERT to initialize the sentence feature extractor, where the former conducted consistency-regularization between the original sentence and its back-translation generated one, and the latter further introduced the manifold mixup~\cite{verma2019manifold} into text classification. Although these methods may achieve decent performances, we believe that they haven't fully explored the inherent knowledge in a PLM. Our work takes a step further in this direction.

\section{Inherent matching capability of a PLM}\label{sec:motivation}
In this section, we will demonstrate the inherent topic matching capability of BERT which motivates our method. %
Utilizing PLMs for a downstream task has become common since it often brings a significant performance boost~\cite{zhu2020incorporating,chen2020mixtext}. In the context of semi-supervised learning, a PLM is usually employed for initializing the network before performing semi-supervised training. However, the value of a PLM can go beyond a good initial model or feature extractor. In particular, a PLM like BERT has already learned certain topic matching capabilities thanks to its pretext tasks.  For example, BERT uses the next sentence prediction (NSP) as one of its pretext tasks.  In this task, the network is asked to discern if two input sentences are two successive sentences in the original corpus. After training on this task, BERT can implicitly acquire topic matching capability since two successive sentences in a paragraph usually share the same topic.

Fig.~\ref{fig:motivation} shows a concrete investigation of the inherent matching capability of BERT. Following the NSP task, we concatenate the sentence and class semantic-related words $C_k$, e.g., ``sports'', via the format: ``[CLS] sentence [SEP] $C_1$ $\cdots$ $C_k$ $\cdots$ $C_K$ [SEP]''. Then we pass the input sequence to a pre-trained BERT and calculate the attention value of each token with respect to each class name. Specifically, this attention value is calculated by averaging the last layer self-attention values across all heads between a token and the appended word $C_k$
. For better visualization, we use different color to show the class that leads to the largest attention value (indicated by the color brightness).

From the visualization, we can see that BERT can automatically match keywords corresponding to the respective class names. Moreover, we find that if we replace the class names with words under the same topic, i.e., family $\to$ boyfriend, sports $\to$ football, the words related to the ground-truth class can still be attended, as shown in Fig.~\ref{fig:motivation_family} and \ref{fig:motivation_sports}.

Finally, we extract BERT last-layer's feature corresponding to each class word $C_k$ and average features align with sentence tokens, and compare the cosine similarity between them. As histograms shown in Fig.~\ref{fig:motivation_family} and \ref{fig:motivation_sports}, we can find that the correct class leads to the highest matching score, although not always by a large margin. 

\begin{figure*}[t]
\centering
\includegraphics[width=0.8\textwidth]{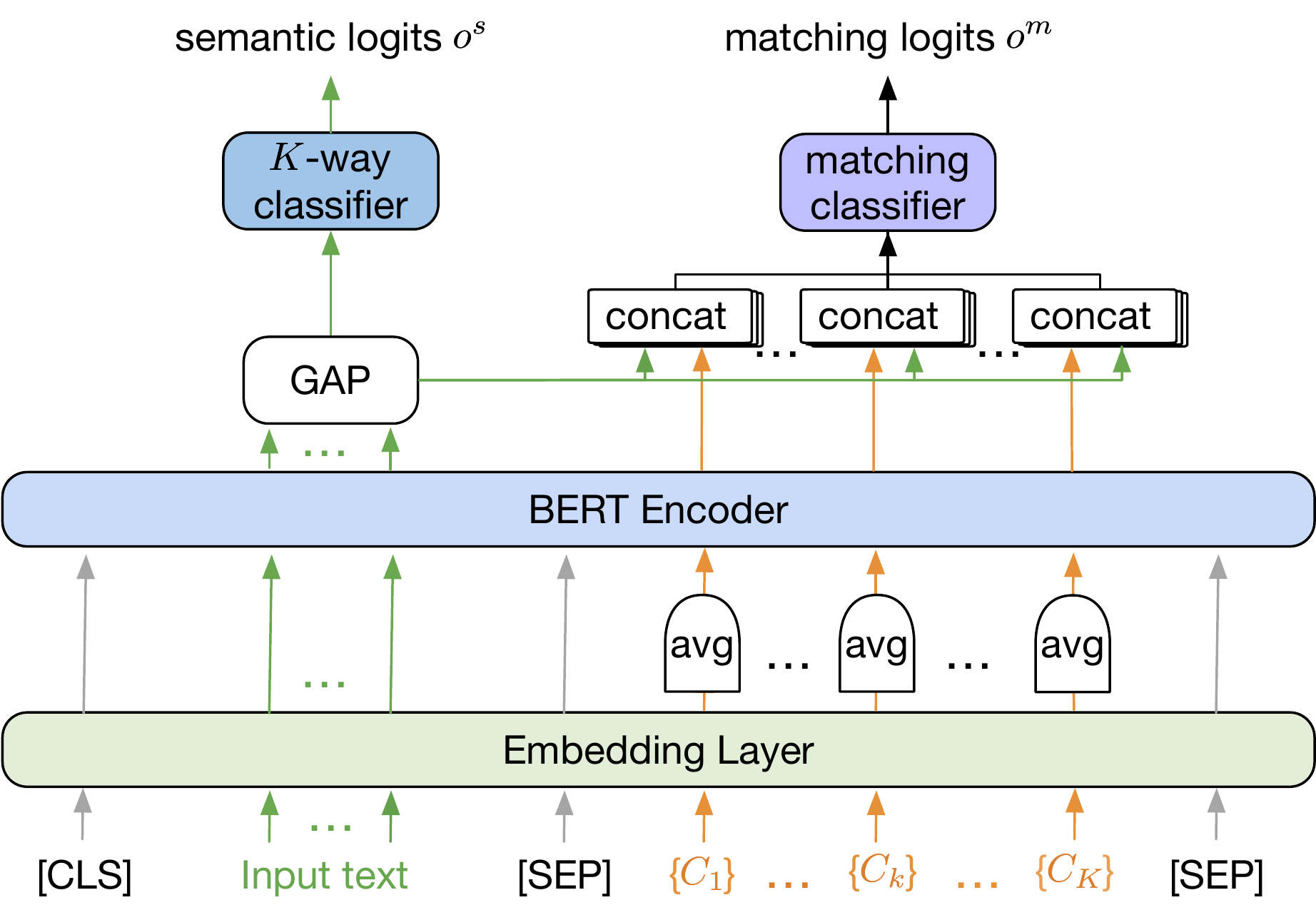}
\caption{Overview of the proposed PCM model. Lines in the same color indicate how the information travels in our model. $\{C_k\}$ denotes the set of class semantic-related words. ``avg'' means the average of word embeddings within the same class. ``GAP'' represents the global average pooling of the input text features. ``concat'' is a feature concatenate operation. We clarify the details of initializing and updating of $\{C_k\}$ in Secs.~\ref{init_cls_semantic_represent} and~\ref{sec:update_components}.
}
\label{fig:pcm_model}
\end{figure*}

\section{Progressive Class-semantic Matching}
To further strengthen the above topic matching capability and use it for classification, we propose to progressively build a sentence-class matching model through the framework of semi-supervised learning. Formally, we aim to build a classifier from a few annotated samples $\mathcal{L} = \{x_1, x_2, \cdots, x_{n_l}\}$, whose labels are $\mathcal{Y}~=~\{ y_1, y_2, \cdots, y_{n_l}\}, y_i \in \{1, \cdots, k, \cdots, K\}$, and a large amount of unlabeled samples $\mathcal{U} = \{x_1, x_2, \cdots, x_{n_u}\}$ (where $n_l \ll n_u$).

The idea is to construct a process that can jointly update three components: (1) 
a standard $K$-way classifier
(2) a matching classifier which matches texts against class semantic representation
(3) the class semantic representation (CSR) itself. The update of each component will help other components and thus can iteratively bootstrap classification performance. We call our method as Progressive Class-semantic Matching (PCM).

\subsection{Three components of PCM}\label{sec:three_components}
Fig.~\ref{fig:pcm_model} shows how we realize the three components. Similar to the example in Section~\ref{sec:motivation}, we construct the input to the BERT by concatenating sentence with class semantic-related words $\{C_i\}, i\in \{1, \cdots, k, \dots, K\}$.  Considering the size of $\{C_i\}$ may vary and the computation cost may increase heavily when the number of classes grows, we calculate an average of 
embeddings of all words belonging to the same class before passing them to the pre-trained BERT encoder. This average embedding is called \textbf{class semantic representation (CSR)}.

The last layer output features corresponding to tokens in the input text are averaged and treated as the sentence representation. On top of the sentence representation, we build a \textbf{standard $K$-way classifier}. We implement it by a two-layer MLP and it will output a set of logits $\{o^s_i\}$ called semantic logits and posterior probabilities $\{p^s_i\}$ after applying Softmax to $\{o^s_i\}$.

In addition to the $K$-way classifier, we also build \textbf{a class-sentence matching classifier} which is realized by another 
MLP applying to the concatenation between the sentence representation and the output features corresponding to each CSR. The output of this matching classifier is called matching logits $\{o^m_i\}$ and Sigmoid function is applied to convert it into the probabilistic form, denoted as $\{p^m_i\}$. Note that the matching classifier is realized in a multi-label formulation, that is, the summation of $\{p^m_i\}$ over all classes is not necessarily equal to 1. It allows the scenario that a sentence matches more than one class and the case that a sentence does not match any class. This design avoids the case that achieving high matching probability for one class merely because its matching score is higher than those of other classes (but it actually with low absolute matching logits for all classes). We empirically find that using this mechanism is helpful for the matching classifier (but not necessarily for the $K$-way predictor as discussed in Section~\ref{abl_study}). 

\subsection{Initialization of CSR}\label{init_cls_semantic_represent}
The proposed PCM model requires an initial CSR, i.e., the average word embedding of a set of class semantic-related words, to start the iteration. 
Although manually choosing a list of seed words (e.g., class names) can be an ideal way for the CSR initialization, it may suffer from leveraging prior knowledge and leads to an unfair comparison to existing SSL algorithms. 
An alternative approach is to
automatically identify a set of class semantic-related words. 
This might be useful for the case that class names in some corpora do not carry a clear semantic meaning, e.g., the rating of reviews.

In this paper, we use the following method to automatically collect the class semantic-related words: 
we start by fine-tuning a pre-trained BERT classifier on the labeled set. Then passing each labeled text into the fine-tuned model and calculate attention values for each token. The attention value of a token is calculated by averaging all the attention received for this token \footnote{Magnitude of the attention value indicates the importance of this token.}. After removing stop words, we retain the top-$j$ e.g., $j=75$, attended words for each class to calculate the initial CSR.

\subsection{Update of three components}\label{sec:update_components}

The three components are progressively updated by seamlessly incorporating them into an SSL framework. In particular, our method is built upon UDA~\cite{xie2019unsupervised}, one of the state-of-the-art approaches in semi-supervised text classification. The idea is to first construct an augmented version of unlabeled data by back translation~\cite{edunov2018understanding} and then enforce the prediction to be consistent through a consistency-regularization loss for unlabeled data. The following describes the detailed updating process:

\smallskip
\noindent\textbf{Update of the standard $K$-way classifier and the class-sentence matching classifier:} 
The update is performed on labeled and unlabeled data at the same time. For labeled data, both classifiers are updated by performing stochastic gradient descent with the following objective function.
\begin{equation}
    \begin{split}
    \mathcal{L}_l & = \frac{1}{n_l}\sum_{j=1}^{n_l} 
    \sum_{i=1}^K  \underbrace{-\mathbb{I}^j_i\log p_i^s(x_j)}_{\text{cross entropy (CE)}} + \\
    & \underbrace{- \mathbb{I}^j_i\log p_i^m(x_j) - (1-\mathbb{I}^j_i)\log \bigl (1-p_i^m(x_j)\bigr )}_{\text{binary cross entropy (BCE)}},    
    \end{split}
\end{equation}
where $p_i^s(x_j)$ and $p_i^m(x_j)$ are the probabilities of $x_j$ belonging to class $i$ from the view of the $K$-way classifier and the matching classifier,  respectively. Since the matching classifier is designed in a multi-label style, we use binary cross-entropy loss for it.
$\mathbb{I}^j_i$ is an indicator whose value equals to 1 if $y_j=i$, and 0 otherwise. 

For unlabeled data, we follow UDA to use a student-teacher alike training strategy, that is, we first use the original sentence input $x_j \in \mathcal{U}$ to obtain the prediction target (similar to a pseudo label) and then enforce the prediction of the back-translated version $x^a_j$ of $x_j$ being close to the prediction target. Formally, if the prediction of one unlabeled sample satisfies all the following rules, the prediction target will be generated:
\begin{equation}\label{eq:3_conditions}
    \begin{cases}
        \text{max}_i\bigl (p_i^s(x_j)\bigr ) >=  \text{confid1} \\
        \text{max}_i\bigl(p_i^m(x_j)\bigr ) >=  \text{confid2} \\
        \text{argmax}_i \bigl (p_i^s(x_j)\bigr ) == \text{argmax}_i \bigl(p_i^m(x_j)\bigr) 
    \end{cases}
\end{equation}
where $\text{confid1}$ and $\text{confid2}$ are two pre-defined confidence thresholds and we empirically find $\text{confid1}=0.95$ and $\text{confid2}=0.7$ performs well in our experiments.
For the $K$-way classifier, the pseudo prediction target is a sharpened posterior probability, i.e., $\hat{p}^s = \text{Softmax}(o^s/T)$ with $T \leq 1$. 
For the matching classifier, we directly generate a pseudo-label by  $\hat{y}_i=\text{argmax}_i\ p_i^m$. The loss function for the unlabeled data is
\begin{align}
    \begin{split}
        \mathcal{L}_u  =  \frac{1}{n_u}\sum_{j=1}^{n_u} & \Bigl( \text{KL}\bigl (p^s(x^a_j),\hat{p}^s(x_j)\bigr ) \\ & + \text{BCE}\bigl (p^m(x^a_j),\hat{y}_i(x_j)\bigr ) \Bigr) 
    \end{split}
\end{align}where $\text{KL}(\cdot,\cdot)$ denotes the KL divergence. 

\begin{table*}[tbp]
    \centering
    \begin{tabular}{c c c c c}
         \toprule
         \textbf{Dataset} & \textbf{Label Type} & \textbf{\# Classes} & \textbf{\# Unlabeled} & \textbf{\# Test}\\
         \hline
         AG News & News Topic & 4 & 20,000 & 7,600 \\
         \hline
         DBpedia & Wikipedia Topic & 14 & 70,000 & 70,000 \\
         \hline
         Yahoo! Answers & QA Topic & 10 & 50,000 & 60,000 \\
         \hline
         IMDB & Review Sentiment & 2 & 10,000 & 25,000 \\
         \bottomrule
    \end{tabular}
    \caption{Statistics of four text datasets.}
    \label{tab:dataset_stats}
\end{table*}
\begin{table*}[t]
    \centering
    \begin{threeparttable}
    \begin{tabular}{c|l | l l l l l}
         \toprule
         \multirow{2}{*}{Dataset} & \multirow{2}{*}{Method} & \multicolumn{5}{c}{Number of Labeled Example Per Clas} \\
         \cmidrule(lr){3-7}
          &  & \multicolumn{1}{c}{\textbf{3}} & \multicolumn{1}{c}{\textbf{5}} & \multicolumn{1}{c}{\textbf{10}} & \multicolumn{1}{c}{\textbf{20}} & \multicolumn{1}{c}{\textbf{50}} \\
         \hline
         \multirow{4}{*}{AG News}
          & BERT-FT & 76.70$\pm$4.72 & 79.90$\pm$2.34 & 83.46$\pm$2.73 & 84.97$\pm$1.73 & 87.35$\pm$0.56 \\
         & UDA & 78.25$\pm$7.61 & 82.97$\pm$2.87 & 86.75$\pm$0.88 & 86.77$\pm$0.10 & 88.23$\pm$0.49 \\
         & MixText & 81.60$\pm$9.04 & 85.84$\pm$1.32 & 85.56$\pm$2.95 & 87.60$\pm$0.48 & 88.14$\pm$0.75 \\
         & \textbf{PCM(ours)} & \textbf{84.85$\pm$0.86} & \textbf{87.20$\pm$0.42} & \textbf{88.31$\pm$0.47} & \textbf{88.34$\pm$0.27} & \textbf{88.85$\pm$0.27} \\
        \Xhline{2\arrayrulewidth}
         \multirow{4}{*}{DBpedia}
         & BERT-FT & 86.68$\pm$2.59 & 91.86$\pm$2.46 & 96.60$\pm$0.46 & 97.84$\pm$0.23 & 98.59$\pm$0.22 \\
         & UDA & 93.51$\pm$2.23 & 95.88$\pm$2.78 & 97.26$\pm$1.50 & 98.59$\pm$0.04 & 98.93$\pm$0.06 \\
         & MixText & 93.25$\pm$0.68 & 96.93$\pm$0.41 & 98.39$\pm$0.09 & 98.64$\pm$0.18 & 98.84$\pm$0.05 \\
         & \textbf{PCM(ours)} & \textbf{94.37$\pm$0.49} & \textbf{97.04$\pm$0.68} & \textbf{98.70$\pm$0.04} & \textbf{98.80$\pm$0.06} & \textbf{99.07$\pm$0.05} \\
        \Xhline{2\arrayrulewidth}
         \multirow{4}{*}{Yahoo!}
         & BERT-FT & 45.93$\pm$3.67 & 50.75$\pm$4.32 & 61.84$\pm$2.37 & 63.89$\pm$0.94 & 67.29$\pm$0.68 \\
         & UDA & 48.30$\pm$11.09 & 57.09$\pm$5.69 & 65.15$\pm$1.54 & 67.76$\pm$0.60 & 69.38$\pm$0.78 \\
         & MixText & 60.27$\pm$4.29 & 65.77$\pm$1.78 & 67.23$\pm$1.97 & 68.19$\pm$1.33 & 69.11$\pm$0.73 \\
         & \textbf{PCM(ours)} & \textbf{63.52$\pm$2.63} & \textbf{67.09$\pm$0.54} & \textbf{68.34$\pm$1.03} & \textbf{69.21$\pm$0.42} & \textbf{70.28$\pm$0.47} \\
        \Xhline{2\arrayrulewidth}
         \multirow{4}{*}{IMDB}
         & BERT-FT & 60.11$\pm$2.41 & 65.17$\pm$8.39 & 73.20$\pm$2.97 & 78.70$\pm$6.75$^\dag$ & 83.91$\pm$1.13 \\
         & UDA & 63.01$\pm$1.07 & 71.90$\pm$10.80 & 89.05$\pm$1.70 & 90.20$\pm$0.54$^\ddag$ & 90.41$\pm$0.45 \\
         & MixText & 56.27$\pm$3.46 & 71.89$\pm$4.89 & 83.38$\pm$3.35 & 86.27$\pm$1.36 & 88.30$\pm$1.24 \\
         & \textbf{PCM(ours)} & \textbf{73.86$\pm$1.04} & \textbf{86.06$\pm$0.74} & \textbf{89.94$\pm$0.44} & \textbf{91.10$\pm$0.28} & \textbf{91.15$\pm$0.15} \\
         \bottomrule
    \end{tabular}
    \caption{Test accuracy (\%) of all comparing methods on four datasets. Models are trained with 3/5/10/20/50 labeled data per class. $\pm$ denotes the Standard Error of the Mean (S.E.M.) over three random sampled label sets. Best results are indicated as bold.
    }
    \label{tab:main_results}
    \begin{tablenotes}
        \small
        \item $^\dag$ Single run accuracy (81.6\%) is reported in UDA~\cite{xie2019unsupervised} for a reference. $^\ddag$ This number is reasonable on one GPU card with 11GB memory. See experimental tutorial~\cite{uda2019exptutorial} for details.
	\end{tablenotes}
	\end{threeparttable}
\end{table*}

\smallskip
\noindent\textbf{Update of CSR:} The initialized CSR might not be accurate or comprehensive enough to represent the class semantics. 
Similar to the approach proposed in Section~\ref{init_cls_semantic_represent}, 
we use the newly updated model to collect a better CSR. The collection process on labeled sentences is still as described in Section~\ref{init_cls_semantic_represent}. While the same extraction operation is performed on unlabeled texts only when they satisfy the conditions in Eq.~\ref{eq:3_conditions}. 
We update the CSR whenever the number of validation set\footnote{Please note that we do not use any label information here.} samples meeting conditions in Eq.~\ref{eq:3_conditions} increases.
Generally, during the course of semi-supervised learning, the classifiers become stronger and the selected class-related words tend to become more accurate. Table~\ref{tab:supporting_words_list} gives an example to show the difference of most attended words between initialization and after training.

\section{Experimental results}
In this section, we perform the experimental study of the PCM method on four text datasets.

\smallskip
\noindent\textbf{Datasets}
Following MixText~\cite{chen2020mixtext}, we use four datasets, namely, AG News~\cite{zhang2015character}, DBpedia~\cite{lehmann2015dbpedia}, Yahoo! Answers~\cite{chang2008importance}, and IMDB~\cite{maas2011learning} for our experiments.  We use the same data splits as in MixText~\cite{chen2020mixtext}. 
The detailed statistics of the four datasets are presented in Table~\ref{tab:dataset_stats}.

\smallskip
\noindent\textbf{Implementation details}
Same as MixText~\footnote{\url{https://github.com/GT-SALT/MixText} (2-clause BSD License)}, we use back-translation to perform data augmentation. Two languages, German and Russian, are chosen as the intermediate language. The back-translation texts on Yahoo! Answers are provided by MixText, and we directly use them. For the other three datasets, we generate the back-translation data by ourselves (with Fairseq toolkit~\cite{ott2019fairseq}).

We use the input format ``[CLS] Sentence [SEP]'' for all the baseline methods. We empirically find this format leads to the overall best performance. Meanwhile, this format actually brings performance improvement to both UDA and MixText methods. So we are comparing against stronger baselines in our paper. 

Due to BERT's length limit, we only kept the last 256 tokens for IMDB and the first 256 tokens for the other datasets during training. We use the same learning-rate setting for all methods: 5e-6 for the BERT encoder and 5e-4 for the classifier (i.e., a two-layer MLP with a 128 hidden size and $tanh$ as its activation function). All our experiments were run on a GeForce RTX 2080 Ti GPU and each experiment takes around 5 hours.

\smallskip
\noindent\textbf{Comparing methods}
We compare the proposed PCM method with three baselines: (1) fine-tuning the pre-trained BERT-based-uncased model on the labeled texts directly, denote as \textbf{BERT-FT}. (2) Unsupervised data augmentation method (UDA)~\cite{xie2019unsupervised} and (3) the recently proposed MixText method \cite{chen2020mixtext}. To make a fair comparison, we conduct all experiments based on the same codebase released by the authors of MixText~\cite{chen2020mixtext}.

\begin{figure}
\centering
\includegraphics[width=0.9\textwidth]{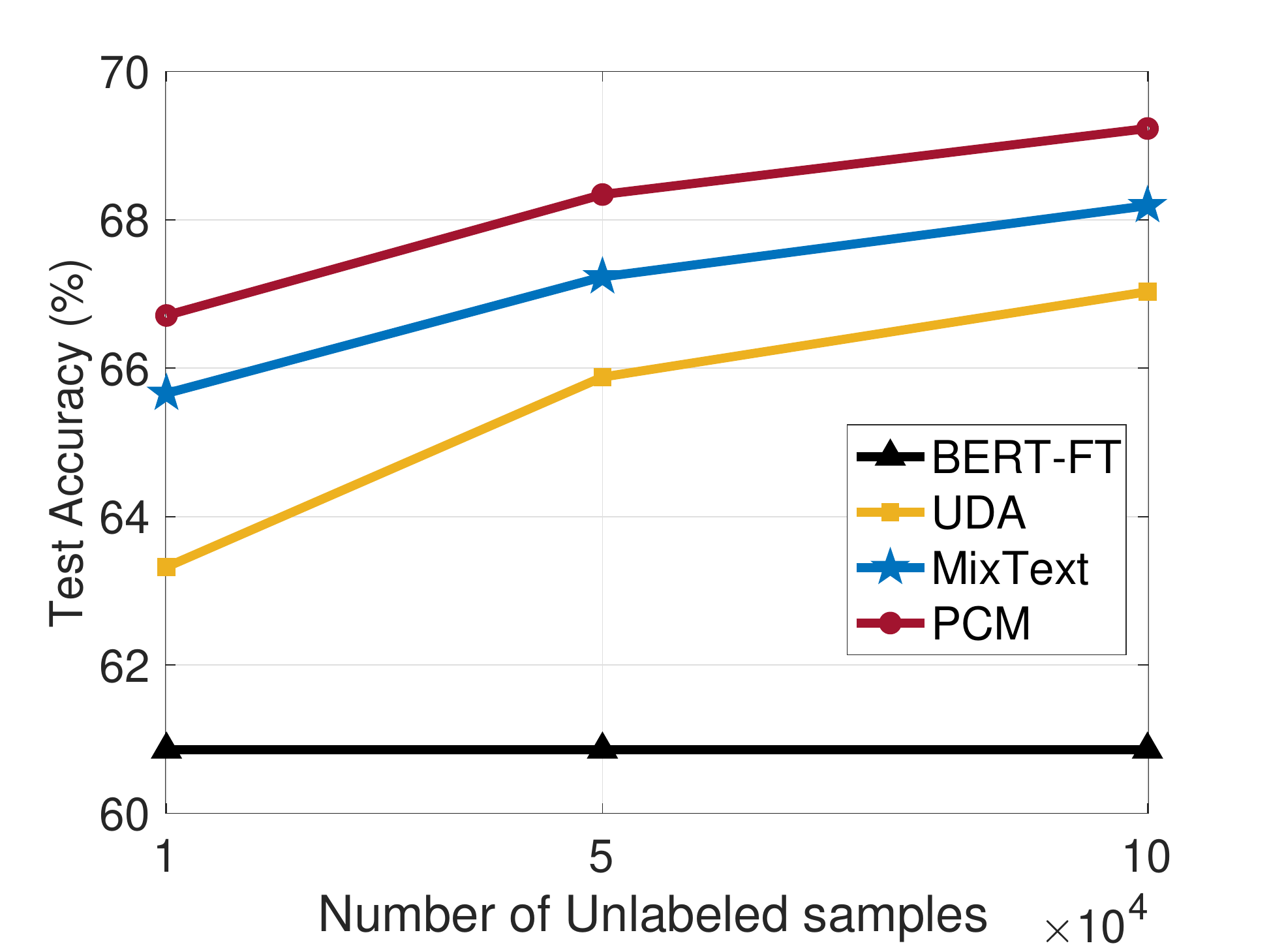}
\caption{Accuracy on varying number of unlab. data.}
\label{fig:abl_more_ul}
\end{figure}

\subsection{Main results}

Table~\ref{tab:main_results} presents the performance comparison of the proposed PCM method and other baselines on different datasets. From that, we can have the following observations. (1) By using BERT, all methods achieve reasonable performance. Even the BERT fine-tune baseline achieves good performance when there are ten samples per class. However, BERT fine-tune is still inferior to the semi-supervised approaches, especially when the number of training samples becomes smaller or the classification task becomes more challenging. (2) As expected, the MixText method excels than UDA in most cases, but performs similarly when the number of labeled samples becomes large (e.g., 50 labels/class). Since the proposed method could also be incorporated into MixText, it might be able to boost its performance. 
(3) 
the proposed PCM methods
achieves significant performance improvement over UDA approaches. Please note that PCM is built on top of the UDA method and this performance gain indicates the effectiveness of using the proposed progressive training process. 
(4) 
It is clear that PCM can not only \textbf{always outperform other baselines} and achieve state-of-the-art text classification performance on all four datasets, but also have \textbf{smaller standard error and be more stable}. PCM performs especially well when the number of labeled samples becomes small. A much larger performance gain is observed when only three labeled samples are available. 

Furthermore, we compare PCM to baselines with 10 labeled data per class and varying number of unlabeled ones on Yahoo! Answers dataset (range from 10,000 to 100,000 unlabeled samples). Fig.~\ref{fig:abl_more_ul} shows that PCM continuously benefits from more unlabeled data and can be consistently superior than other methods.

\begin{figure}
\centering
\includegraphics[width=0.9\textwidth]{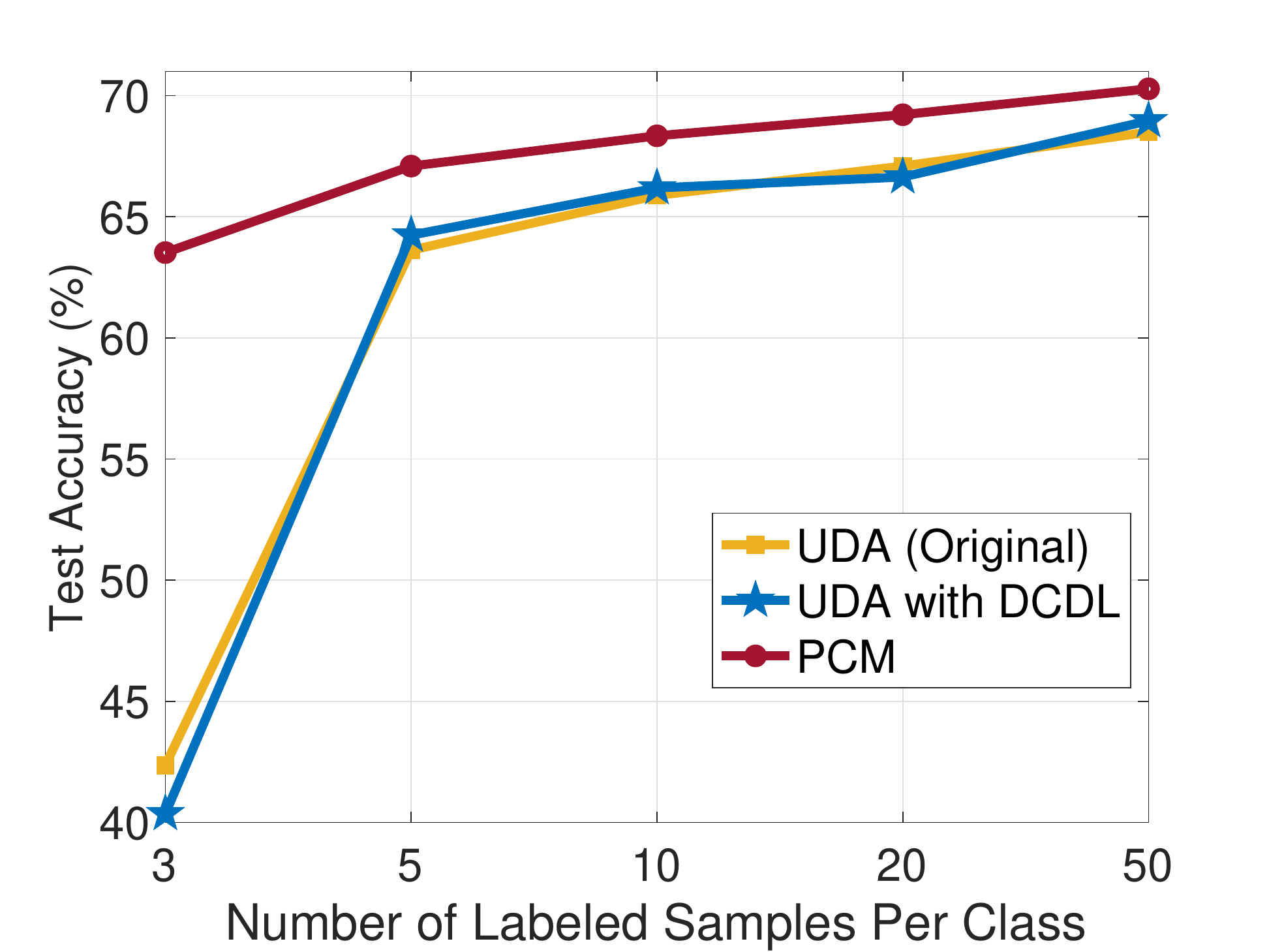}
\caption{Ablation study on the DCDL strategy in PCM.}
\label{fig:abl_uda_bce}
\end{figure}

\subsection{Ablation studies}\label{abl_study}
PCM model consists of several components. In this section, we perform ablation studies to examine their impact. Most of these studies are performed on Yahoo! Answers dataset with one identical labeled set, unless otherwise specified.

\begin{table*}[tbp]\footnotesize
    \centering
    \begin{tabularx}{1.0\textwidth}{lX}
         \toprule
         Initial & bush, car, bomb, killed, chancellor, black, moscow, inter, leftlist, putin, story, presidential, texas, president, campaign, documents, ap, unearthed, caracas, ... \\
         \hline
         Final & iraq, president, iraqi, government, baghdad, military, palestinian, security, nuclear, prime, minister, country, israeli, leader, war, peace, gaza, iran, israel, troops, ... \\
         \bottomrule
    \end{tabularx}
    \caption{The class semantic-related word lists on class ``world'' of AG News dataset. The top row is the initial class semantic-related words obtained from fine-tuned BERT, while the bottom one is the final class semantic-related words after PCM training with upper initial words. All models are trained on the 3 labels per class case.}
    \label{tab:supporting_words_list}
\end{table*}

\begin{table}[t]\footnotesize
    \centering
    \begin{tabular}{c c| c c c c c}
         \toprule
         \multirow{2}{*}{\textbf{$p^s$}} & \multirow{2}{*}{\textbf{$p^m$}} & \multicolumn{5}{c}{Label Number Per Class} \\
         \cmidrule(lr){3-7}
          &  & \multicolumn{1}{c}{\textbf{3}} & \multicolumn{1}{c}{\textbf{5}} & \multicolumn{1}{c}{\textbf{10}} & \multicolumn{1}{c}{\textbf{20}} & \multicolumn{1}{c}{\textbf{50}} \\

         \hline
         \xmark & \cmark & 10.01 & 10.45 & 10.01 & 10.21 & 10.05 \\
         \cmark & \xmark & 49.51 & 65.32 & 65.70 & 67.88 & 68.43 \\
         \cmark & \cmark & \textbf{63.52} & \textbf{67.09} & \textbf{68.34} & \textbf{69.21} & \textbf{70.28} \\
         \bottomrule
    \end{tabular}
    \caption{Ablation study on the importance of two classifiers of the proposed PCM model.}
    \label{tab:abl_structure}
\end{table}

\begin{table}[t]\footnotesize 
    \centering
    \begin{tabular}{c|c c c c c}
         \toprule
         \multirow{2}{*}{update CSR} & \multicolumn{5}{c}{Number of Labeled Example Per Class} \\
         \cmidrule(lr){2-6}
          & \multicolumn{1}{c}{\textbf{3}} & \multicolumn{1}{c}{\textbf{5}} & \multicolumn{1}{c}{\textbf{10}} & \multicolumn{1}{c}{\textbf{20}} & \multicolumn{1}{c}{\textbf{50}} \\

         \hline
         \xmark & 39.49 & 66.04 & 66.41 & 67.09 & 68.85 \\
         \cmark & \textbf{63.52} & \textbf{67.09} & \textbf{68.34} & \textbf{69.21} & \textbf{70.28} \\
         \bottomrule
    \end{tabular}
    \caption{Ablation study on the importance of updating the CSR during training of PCM.}
    \label{tab:update_suporting_words}
\end{table}

\smallskip
\noindent\textbf{1. The importance of using two classifiers in PCM.}
The proposed PCM model contains a $K$-way classifier (i.e., $p^s$) and a matching classifier (i.e., $p^m$), and they are jointly trained in the proposed process. We investigate the role of them by constructing a variant of PCM by only using either one of them. As the results shown in Table~\ref{tab:abl_structure}, without using the $K$-way classifier, the method totally fails to a random guess. In contrast, only keeping the $K$-way classifier can obtain reasonable results. More interestingly, this variant actually performs better than UDA on 3 and 5 label cases (see the Table~\ref{tab:main_results}). The difference between this variant and UDA is that the former appends CSR to the input sequence. Its good performance shows that merely appending CSR can be helpful for semi-supervised text classification. Finally, we can see that using both classifiers can lead to the best performance. This clearly validates the necessity of the proposed joint learning process.

\smallskip
\noindent \textbf{2. If using the dual-classifier-dual-loss is the key to success?}
In our method, we utilize a slightly unconventional dual-classifier-dual-loss strategy (DCDL): the pseudo-labels are generated by checking the agreement of the two classifiers, and two losses, i.e., BCE and CE, are used for training those two classifiers. One may suspect that our good performance actually stems from this DCDL scheme rather than leveraging BERT's matching capability. To investigate this problem, we conduct an ablation study by modifying UDA with this strategy. Specifically, we use two classifiers, one trained from the BCE loss and the other one trained from the CE loss. The pseudo-prediction targets are generated by following the same strategy as in PCM. The result is shown in Fig.~\ref{fig:abl_uda_bce}. As seen, simply incorporating this training strategy does not necessarily bring better classification accuracy. This result provides evidence that the PCM's good performance can not be simply attributed to the DCDL strategy.

\begin{figure}
\centering
\includegraphics[width=0.9\textwidth]{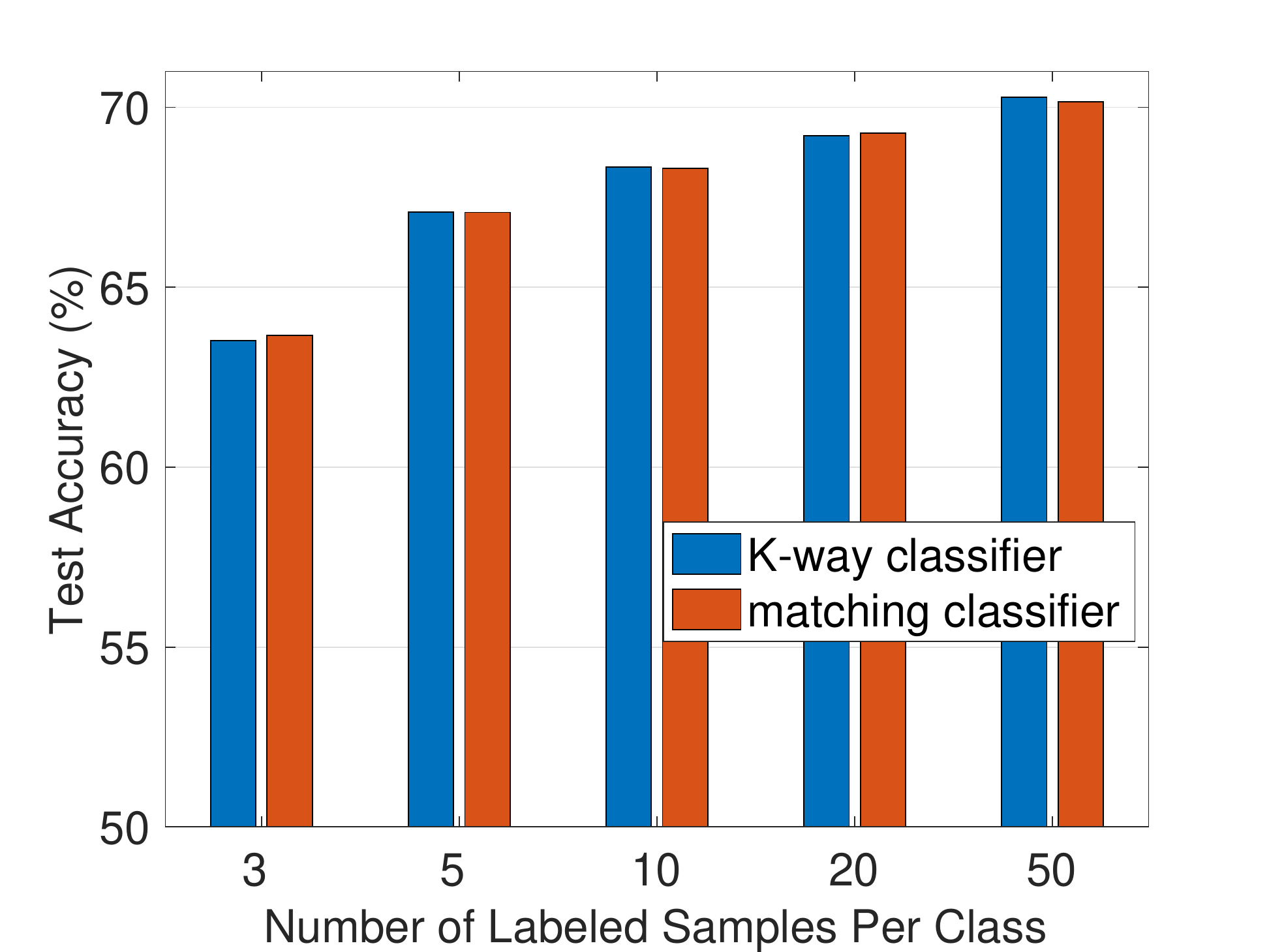}
\caption{Ablation study on classifier quality of PCM.
}
\label{fig:abl_semantic_vs_matching_classifier}
\end{figure}

\smallskip
\noindent \textbf{3. The prediction quality of the $K$-way classifier and the matching classifier.} In our PCM model, the $K$-way classifier is chosen for the final testing phase. We further validate the quality of the matching classifier. As the results presented in Fig.~\ref{fig:abl_semantic_vs_matching_classifier}, the matching classifier gains comparable performance to the $K$-way one. This proves that the collaborative training of two classifiers bootstraps each other to have good prediction capability.

\smallskip
\noindent \textbf{4. The impact of updating CSR.}
Our PCM method dynamically updates the CSR through the training process. In this part, we investigate the impact of this updating process. Table~\ref{tab:update_suporting_words} compares the results obtained by updating or not updating CSR. As seen, updating CSR leads to overall better performance. The difference becomes quite significant when only three labeled samples are used. For example, PCM may fail when the class semantic representation is fixed in the 3-label case.

\section{Limitations and Potential Risks}
One underlying assumption about our findings is that we mainly consider BERT-style pre-trained language models for semi-supervised text classification. The utilization of inherent knowledge of other language models (e.g., GPT~\cite{radford2018improving} and XLNet~\cite{DBLP:journals/corr/abs-1906-08237}) are not explored in this paper and is left for future work.

PCM algorithm has been verified to be effectiveness on texts in English, whether other languages can achieve the same performance improvement is at risk and will be explored in the future.

\section{Conclusion}
In this paper, we proposed a semi-supervised text classification approach by leveraging the inherent topic matching capability in pre-trained language models. The method progressively updates three components, a $K$-way classifier, the class semantic representation, and a matching classifier that matches input text against the class semantic representation. We show that the updating of the three components can benefit each other and achieve superior semi-supervised learning performance.%

\section{Ethics}
In terms of ethics, we do not see immediate concerns for the models we introduce and to the best of our knowledge no datasets were used that have known ethical issues.

\normalem
\bibliography{anthology,custom}
\bibliographystyle{acl_natbib}




\end{document}